# Image-Based Vehicle Classification by Synergizing Features from Supervised and Self-Supervised Learning Paradigms

Shihan Ma and Jidong J. Yang

Smart Mobility and Infrastructure Laboratory
School of Environmental, Civil, Agricultural, and Mechanical Engineering
College of Engineering, University of Georgia, Athens GA, USA

**Abstract**

This paper introduces a novel approach to leverage features learned from both supervised and self-supervised paradigms, to improve image classification tasks, specifically for vehicle classification. Two state-of-the-art self-supervised learning methods, DINO and data2vec, were evaluated and compared for their representation learning of vehicle images. The former contrasts local and global views while the latter uses masked prediction on multi-layered representations. In the latter case, supervised learning is employed to finetune a pretrained YOLOR object detector for detecting vehicle wheels, from which definitive wheel positional features are retrieved. The representations learned from these self-supervised learning methods were combined with the wheel positional features for the vehicle classification task. Particularly, a random wheel masking strategy was utilized to finetune the previously learned representations in harmony with the wheel positional features during the training of the classifier. Our experiments show that the data2vec-distilled representations, which are consistent with our wheel masking strategy, outperformed the DINO counterpart, resulting in a celebrated Top-1 classification accuracy of 97.2% for classifying the 13 vehicle classes defined by the Federal Highway Administration.

## 1. Introduction

Vehicle classification is crucial for highway infrastructure planning and design. In practice, a large quantity of sensors has been installed in state highway networks to collect vehicle information, such as weight, speed, class, and count of vehicles [1]. Many studies have been conducted to classify vehicle types based on sensor data. For example, Wu et al. (2019) used roadside LiDAR data for vehicle classification [2]. The study evaluated traditional machine learning methods (e.g., naive Bayes, k-nearest neighbors (KNN), random forest (RF), and support vector machine) for classifying eight vehicle categories and resulted in a best accuracy of 91.98%. In another study, Sarikan et al. (2017) employed KNN and decision trees for automated vehicle classification, where the inputs were extracted features from vehicle images. The method could distinguish all sedans and motorcycles in the test dataset [3]. Recent developments in vision-based deep learning, inspired by AlexNet [4], have made image-based vehicle classification a popular approach and continue to elevate the image classification benchmark. Zhou et al. (2016) demonstrated a 99.5% accuracy in distinguishing cars and vans and a 97.36% accuracy in distinguishing among sedans, vans, and taxis [5]. Similarly, Han et al. used YOLOv2 to extract vehicle images from videos and applied an autoencoder-based layer-wise unsupervised pretraining to a convolutional neural network (CNN) for classifying motorcycles, transporter vehicles, passenger vehicles, and others [6]. ResNet-based vehicle classification and localization methods were developed using real traffic surveillance recordings, containing 11 vehicle categories, and it obtained a 97.95% classification accuracy and 79.24% mean average precision (mAP) for the vehicle localization task [7]. To ensure the robustness of the models against weather and illumination variation, Butt et al. expanded a large dataset with six common vehicle classes considering adverse illuminous conditions and used it to finetune several pretrained CNN models (AlexNet, GoogleNet, Inception-v3, VGG, and ResNet) [8]. Among those, the finetuned ResNet was able to achieve 99.68% test accuracy. Regardless of the recent success in image-based vehicle classification, most of the models have been developed based on common vehicle categories that are not consistent with the vehicle classes established for engineering practice, such as the Federal Highway Administration (FHWA) vehicle classification, which defined 13 vehicle classes [9] with key axle information, as summarized in Table 1. The illustrative vehicle picture for each class can be found in [10].

Table 1. This is a table. FHWA vehicle classification definitions

| Vehicle Class | Class Includes | Number of Axles | Vehicle Class | Class Includes | Number of Axles |
|---|---|---|---|---|---|
| 1 | Motorcycles | 2 | 8 | Four or fewer axle single-trailer trucks | 3 or 4 |
| 2 | All cars<br>Cars with one- and two- axle trailers | 2, 3, or 4 | 9 | Five-axle single-trailer trucks | 5 |
| 3 | Pick-ups and vans<br>Pick-ups and vans with one- and two-axle trailers | 2, 3, or 4 | 10 | Six or more axle single-trailer trucks | 6 or more |
| 4 | Buses | 2 or 3 | 11 | Five or fewer axle multi-trailer trucks | 4 or 5 |
| 5 | Two-Axle, six-Tire, single-unit trucks | 2 | 12 | Six-axle multi-trailer trucks | 6 |
| 6 | Three-axle single-unit trucks | 3 | 13 | Seven or more axle multi-trailer trucks | 7 or more |
| 7 | Four or more axle single-unit trucks | 4 or more | | | |

Nonetheless, several vehicle classification studies have been conducted with respect to FHWA vehicle classes by focusing on truck classes. Given the detailed axle-based classification rules established by the FHWA, many researchers have used them explicitly for truck classification [11]. In statewide practice, weigh-in-motion (WIM) systems and advanced inductive loop detectors are typically utilized to collect data for truck classification based on the FHWA definition, from which high correct classification rates have been reported for both single-unit trucks and multi-unit trucks [12]. As mentioned previously, besides the traditional sensing technologies, vision-based models have recently been applied in truck classification. Similarly to [5], YOLO was adopted for truck detection. Then, CNNs were used to extract features of the truck components, such as truck size, trailers, and wheels, followed by decision trees to classify the trucks into three groups [13]. This work was continued by further introducing three discriminating features (shape, texture, and semantic information) to better identify the trailer types [14].

As noted in [14], some of the classes in the FHWA scheme only have subtle differences, and deep learning models have potential overfitting issues with their imbalanced datasets. It remains a challenge for vision-based models to successfully classify all 13 FHWA vehicle classes. The objective of this study is to leverage general representations distilled from the state-of-the-art self-supervised methods (DINO [15] and data2vec [16]) as well as specific wheel positional features extracted by YOLOR [17] to improve vehicle classification. Our results show that vehicle representations are the primary features for classifying different vehicle types while wheel positions are complementary features to help better distinguish similar vehicle classes, such as classes 8 and 9, where the only salient feature difference between them is the number of axles. To reinforce this feature complementarity, the general vehicle representations from self-supervised methods were further finetuned in a subsequent supervised classification task together with a random wheel masking strategy, which is compatible with the contextualized latent representations distilled by the data2vec method [16]. As a result, our method significantly improves the classification performance and achieved a Top-1 accuracy of 97.2% for classifying the 13 FHWA vehicle classes. This paper is organized into five sections. Section 2 describes the dataset, followed by our proposed method in Section 3, experiments in Section 4, and finally conclusions and discussions in Section 5.

## 2. Data Description

The dataset contains 7898 vehicle images collected from two sources: the Georgia Department of Transportation (GDOT) WIM sites and the ImageNet [18] opensource dataset. The GDOT data were collected by the cameras installed at selected WIM stations. A total of 6571 vehicles images was collected from the GDOT WIM sites, consisting mainly of the common classes, such as class 2 and class 9. The number of images across the 13 vehicle classes was not well balanced. Rare classes in the GDOT image dataset, such as classes 1, 4, 7, 10, and 13, contained a small number of images. In compensating for these low-frequency classes, an additional 1327 vehicle images were extracted from the ImageNet. Figure 1 summarizes the distribution of vehicle images across the 13 FHWA vehicle classes from both sources, the WIM sites and ImageNet. Exemplar images from each data source are shown in Figure 2. Given the varying scale of vehicles relative to the image frame, the vehicles were cropped from the original images to remove the irrelevant background information. All models were trained based on the cropped vehicle images.

It should be noted that the number of vehicle axles is considered an important feature in FHWA vehicle class definition. For instance, class 8, class 9, and class 10 are both one-trailer trucks. The only difference among these classes is the number of axles (represented by the number of wheels in vehicle images). In addition, the relative locations of wheels are also different across vehicle categories. To extract these particular wheel positional features, a wheel detector was trained to locate all wheels in an image, as shown in Figure 3. With the locations of wheels being identified, the relative positions of all wheels were computed by dividing the wheel spacings ($D_i$) by the maximum distance (i.e., distance between the center of the leftmost wheel and the center of the rightmost wheel) as depicted in Figure 4. This normalization process exists to remove the effect of different camera angles and unifying wheel positional information from different vehicle sizes and scales. The resulting normalized wheel positional features were a vector of the relative wheel positions, which were used to complement the features extracted by the vision transformer models for the downstream classification task.

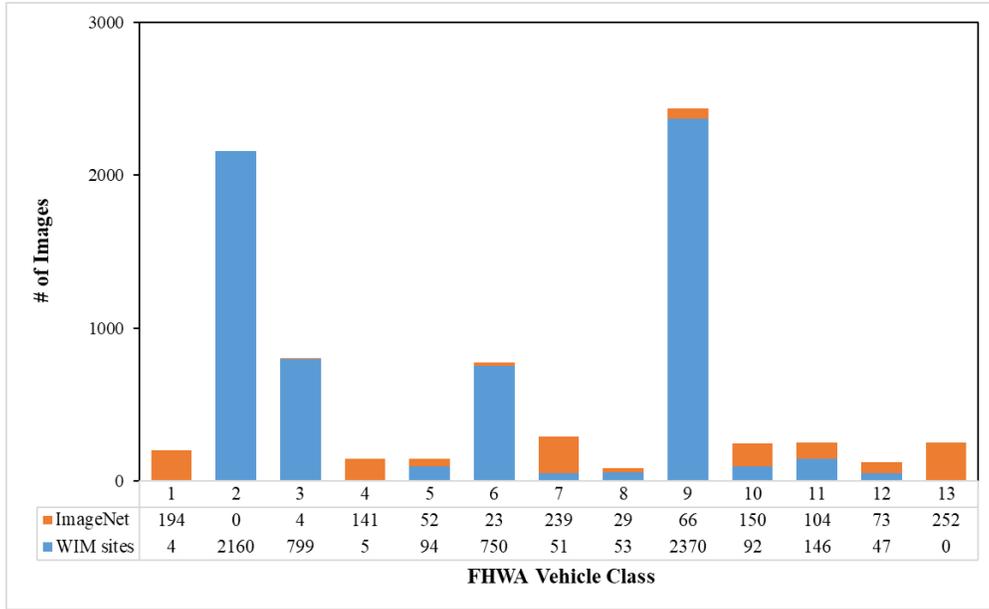

Figure 1. Distribution of image data among 13 FHWA classes from WIM sites and ImageNet.

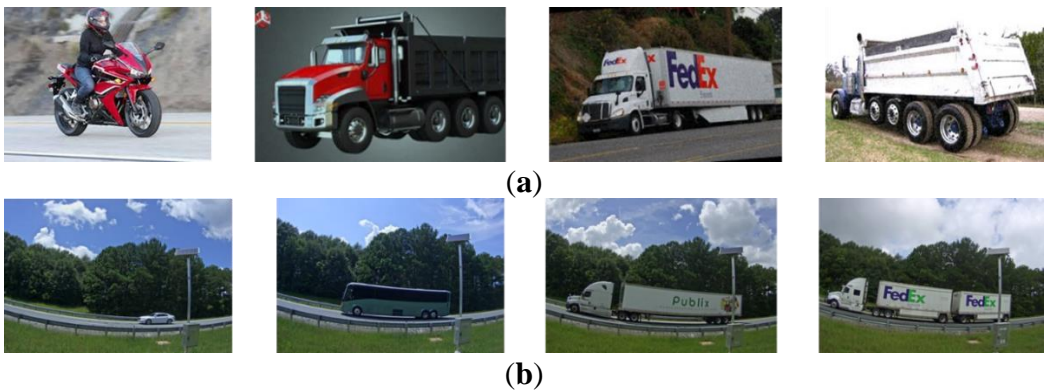

Figure 2. Examples of vehicle images in the dataset. (a) Vehicle images from the ImageNet. (b) Vehicle images collected at the GDOT WIM sites.

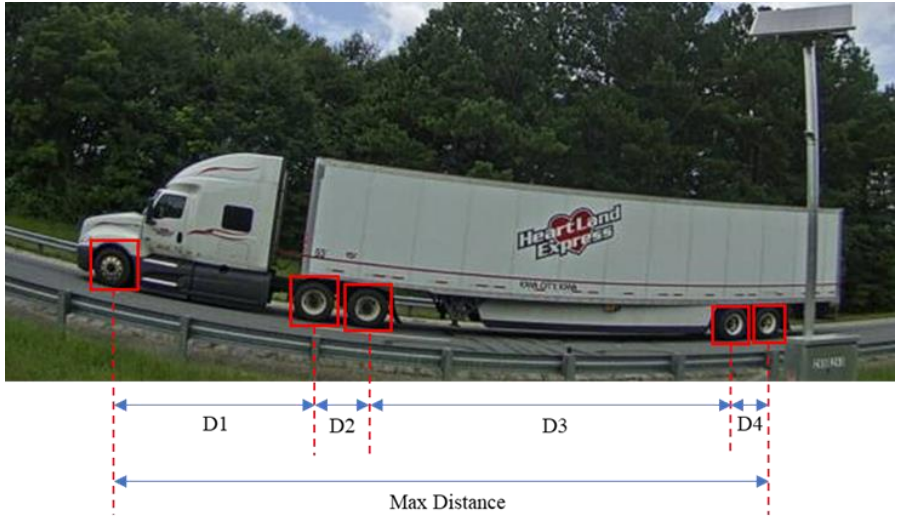

Figure 3. Illustration of wheel positional feature extraction (D1, D2, D3, and D4 are the center-to-center horizontal distances between two successive wheels from left to right).

## 3. Method

Artificial intelligence, especially deep learning, has grown dramatically over the past decade, fulfilling many real-world necessities. Many creative and influential models have been introduced especially in the cognitive computing, computer vision (CV), and natural language processing (NLP) areas. Some models have accomplished multidisciplinary success. One typical example is the transformer [19], which was first developed for NLP and has been successfully applied to vision tasks. The original transformer consists of multiple layers of encoder/decoder blocks, each of which has a combination of three key modules: a self-attention module, feedforward network modules, and layer normalization. Given its increasing popularity, the self-attention mechanism and adapted ViT architectures have been widely adopted across different fields (e.g., [20,21]). In this study, we leveraged the pretrained ViT encoders with the state-of-the-art self-supervised learning methods (i.e., DINO and data2vec) and complemented the ViT representations with wheel positional features retrieved from a finetuned object detection model (i.e., YOLOR). The two sets of features (i.e., ViT representations and wheel positional features) were harmonized by a wheel masking strategy during the classifier training. Our proposed method has shown dramatically improved classification performance when data2vec is used as the pretraining method and the ViT encoder is finetuned during the subsequent classifier training stage.

### 3.1. Vision Transformer

The transformer's architecture has recently been adapted to successfully handle vision tasks [22]. The ViT model has been demonstrated to achieve comparable or better image classification results than traditional CNNs [23–25]. Specifically, ViT leverages embeddings from the transformer encoder for image classification. As depicted in Figure 4, the input image is first divided into small image patches. Each patch is flattened and linearly projected to a latent vector dimension, which is then kept constant throughout all layers. The latent vector is learnable and referred to as patch embedding. A positional embedding is added to the patch embedding process to retain the spatial relationship among the image patches. The positional embedding process is illustrated using $2 \times 2$ patches in Figure 4. In practice, usually $7 \times 7$ or more patches are used. A class token is added and serves as a learnable embedding to the sequence of embedded patches. The learned representations from the encoder are passed to a multi-layer perception (MLP) for image classification. ViT has surpassed many popular CNN-based vision models, such as Resnet152 [22].

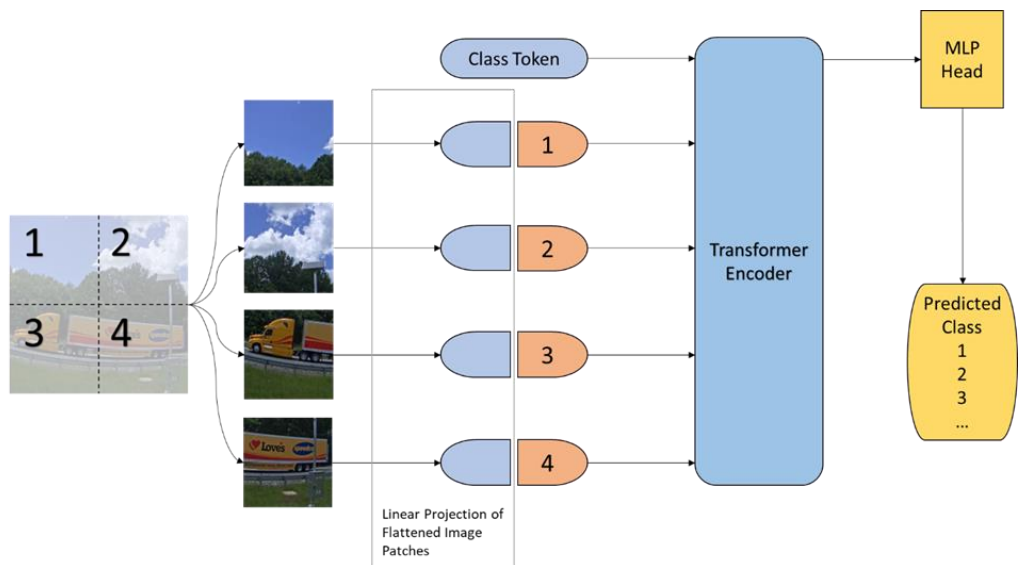

Figure 4. Illustration of a vision transformer.

### 3.2. Self-Supervised Pretraining

To leverage the large number of unlabeled images (e.g., ImageNet [18]), self-supervised learning is adopted for pretraining a ViT encoder (base network). Two state-of-the-art methods: (1) self-distillation with no label (DINO) [15] and (2) data2vec [16] were evaluated in this setting. Figure 5 illustrates the structure of DINO, where input images are randomly cropped to form different views (global and local views) and fed to a teacher ViT and a student ViT, respectively. Therefore, only global view images are passed to the teacher ViT while both global and local view images are passed to the student ViT. In this way, the student model is able to extract multi-scale features. With the encodings from both ViTs, a "softmax" function is applied to produce two probability distributions, $p_1$ and $p_2$. A cross-entropy loss is then computed between p1 and p2. During the training, the parameters of the student ViT are updated by a stochastic gradient descent, while the parameters of the teacher ViT are updated from the exponential weighted average of the student ViT's parameters.

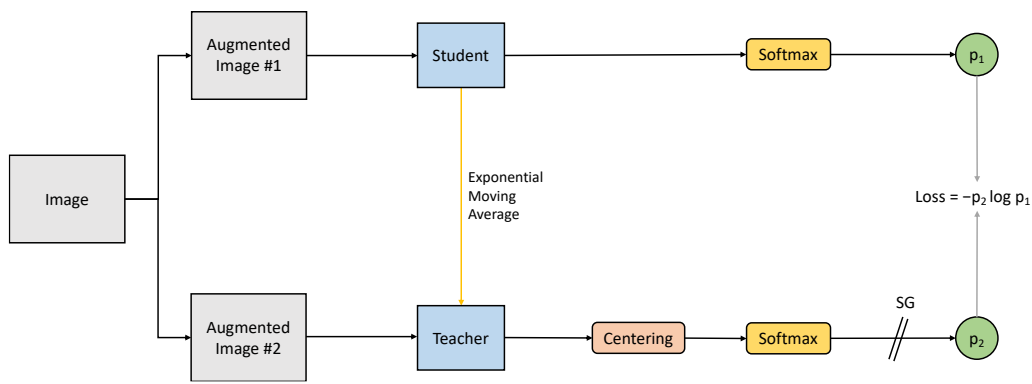

Figure 5. Illustration of DINO.

The data2vec was recently proposed by Baevski et al. [16], which represents a general self-supervised learning framework for speech, NLP, and computer vision tasks. The structure of data2vec is illustrated in Figure 6. Similar to DINO, data2vec also employs a teacher–student paradigm. The teacher generates representations from the original input image, while the student generates representations from the masked image. Different from DINO, data2vec predicts the masked latent representation and regressed multiple neural network layer representations instead of just the top layer. Instead of the cross-entropy loss in DINO, a smooth $L_1$ loss is used in data2vec.

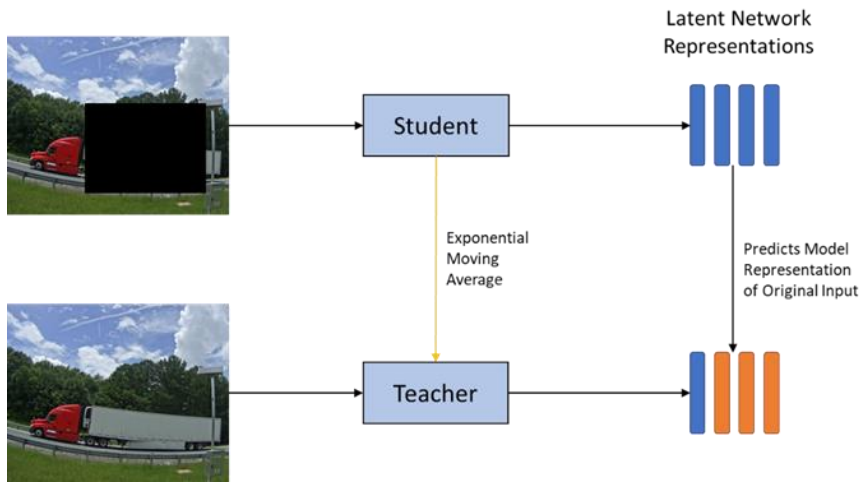

Figure 6. Illustration of data2vec.

### 3.3. Wheel Detection

As discussed previously, the relative wheel positions are critical features according to the FHWA vehicle classifications. Therefore, being able to detect all wheels in vehicle images provides more definitive features for vehicle classification. To leverage the existing object detection models, three real-time object detection architectures, i.e., Faster R-CNN [26], YOLOv4 [27], and YOLOR [17], were evaluated as potential wheel detectors.

Based on our experiments, both YOLOv4 and YOLOR achieved a mAP of 99.9%, slightly outperforming the Faster R-CNN model (mAPs = 99.0%). In light of the faster inference speed, YOLOR was chosen as the wheel detector in our study for extracting wheel positional features. In our experimental setting, the YOLOR model played dual roles: (1) detecting vehicles (with bounding boxes) so that they could be cropped out for further processing (as mentioned previously, all the models were trained and tested on cropped images rather than the original images) and (2) extracting wheel positional features, which were combined with the ViT features for the vehicle classification task. The fusion of these two sets of features was achieved by the end-to-end training of a composite model architecture as discussed in the following section.

### 3.4. Composite Model Architecture

A composite model architecture is proposed to improve vehicle classification by harnessing the features extracted from both ViT and wheel detection models, as shown in Figure 7. The input image is fed to YOLOR for vehicle and wheel detection. Then, the vehicle image is cropped based on the vehicle bounding boxes output from the YOLOR and resized to 224 × 224, which is the input size for the ViT encoder. Features extracted from the ViT encoder and the wheel positional features are concatenated and fed to a multi-layer perceptron (MLP) to classify the 13 FHWA vehicle classes. The wheel locations detected by YOLOR provide wheel positional features that are complementary to those features extracted by the ViT encoder since the former provides localized details on axle configuration while the latter emphasizes vehicle features at a coarser and larger scale (i.e., not necessarily attending to the wheel position details). To further reinforce this complementarity, one wheel was randomly masked when finetuning the ViT encoder. The vanilla ViT and pretrained ViTs by DINO and data2vec were all evaluated in this composite architecture setting. The experimental results are presented in the Experiments section.

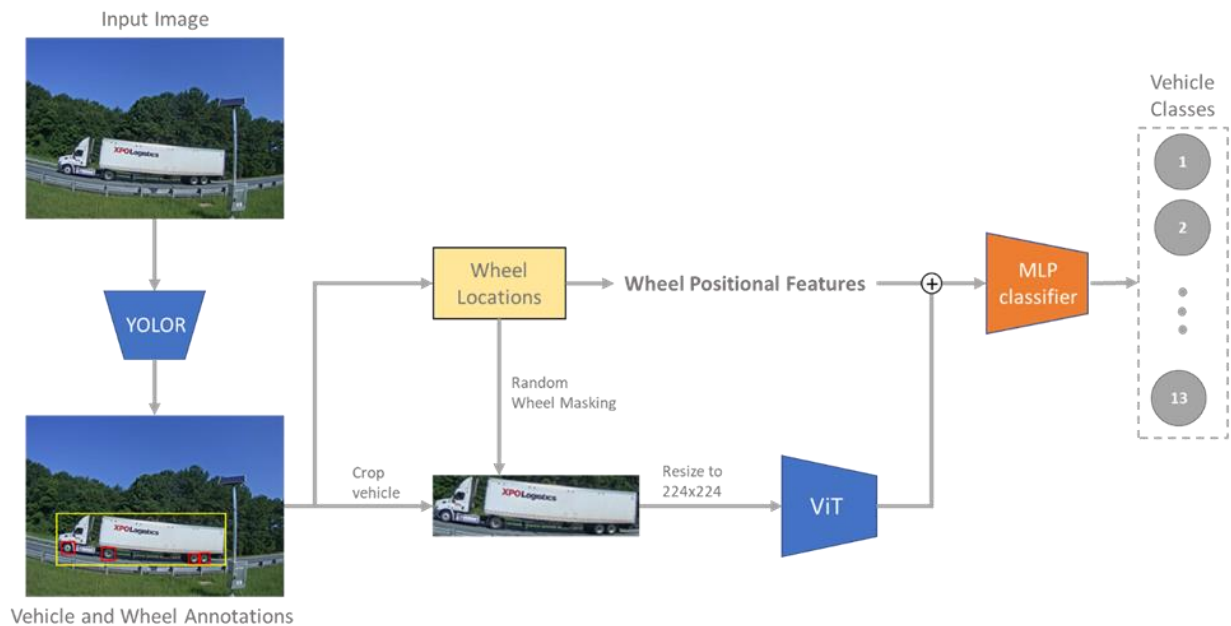

Figure 7. Structure of the composite model architecture.

## 4. Experiments

### 4.1. Effects of Self-Supervised Pretraining

The ViT model was trained with Adam [28] and a batch size of 64. The learning rate was initially set to $6 \times 10^{-5}$. For model development and evaluation, the dataset was split with 80% being relegated to training and 20% to testing. All cropped images were resized to $224 \times 224$ pixels and evenly divided into 14 x 14 patches. Two self-supervised methods, DINO and data2vec, were evaluated as a pretraining stage for the supervised classification task. An ImageNet-pretrained ViT with DINO and data2vec was utilized in this study. Our model training was conducted under two settings: (1) freezing the ViT backbone and only training the MLP classifier and (2) finetuning the ViT backbone while training the MLP classifier. For comparison purposes, the original ViT model was also trained end-to-end in a supervised fashion. The results are summarized in Table 2.

     As shown in Table 2, the Top-1 accuracy, weighted average precision, and weighted average recall from the pretrained ViT models are significantly higher than those from the supervised ViT model regardless of whether the ViT backbone was frozen or not during the classifier training. There was a clear performance boost when the ViT encoder was finetuned during the classifier training. For the finetuned models, the ViT + DINO network performed slightly better than the ViT + data2vec.

**Table 2.** Comparison of classification performance by different ViT training settings.

| Network | Top-1 Acc. (%) | Weighted Avg. Precision (%) | Weighted Avg. Recall (%) |
|---|---|---|---|
| ViT | 90.7 | 90.7 | 90.7 |
| ViT + DINO (freeze-encoder) | 94.6 | 94.6 | 94.6 |
| ViT + DINO | 95.6 | 95.7 | 95.6 |
| ViT + data2vec (freeze-encoder) | 93.5 | 93.5 | 93.5 |
| ViT + data2vec | 95.0 | 95.0 | 95.0 |

For a detailed performance comparison across vehicle classes, the classification reports of the ViT, ViT + DINO, and ViT + data2vec are presented in Table 3. Overall, ViT performed well in the common classes (classes 2, 3, 6, and 9). However, for minority classes (classes 1, 4, 5, 7, 8, 10, 11, 12, and 13), pretrained ViTs reported much better precision, recall, and F1-scores.

**Table 3.** Comparison of classification reports of the ViT model with and without self-supervised pretraining.

|  | ViT without Pretraining | | | ViT Pretrained with DINO | | | ViT Pretrained with data2vec | | |
|---|---|---|---|---|---|---|---|---|---|
|  | Precision (%) | Recall (%) | F1-Score (%) | Precision (%) | Recall (%) | F1-Score (%) | Precision (%) | Recall (%) | F1-Score (%) |
| Class 1 | 88.0 | 55.0 | 67.7 | 100.0 | 100.0 | 100.0 | 100.0 | 100.0 | 100.0 |
| Class 2 | 96.6 | 98.6 | 97.6 | 98.4 | 99.1 | 98.7 | 98.2 | 99.8 | 99.0 |
| Class 3 | 93.5 | 89.4 | 91.4 | 97.5 | 95.7 | 96.6 | 99.4 | 95.0 | 97.1 |
| Class 4 | 73.5 | 86.2 | 79.4 | 100.0 | 93.1 | 96.4 | 96.4 | 93.1 | 94.7 |
| Class 5 | 90.0 | 62.1 | 73.5 | 88.5 | 79.3 | 83.6 | 88.5 | 79.3 | 83.6 |
| Class 6 | 90.3 | 96.1 | 93.1 | 87.2 | 96.8 | 91.7 | 88.2 | 96.1 | 92.0 |
| Class 7 | 66.2 | 74.1 | 69.9 | 95.8 | 79.3 | 86.8 | 84.9 | 77.6 | 81.1 |
| Class 8 | 57.1 | 50.0 | 53.3 | 78.6 | 68.8 | 73.3 | 90.0 | 56.3 | 69.2 |
| Class 9 | 96.2 | 98.2 | 97.2 | 96.8 | 98.4 | 97.6 | 97.0 | 98.4 | 97.7 |
| Class 10 | 64.4 | 60.4 | 62.4 | 90.0 | 75.0 | 81.8 | 81.4 | 72.9 | 76.9 |
| Class 11 | 78.9 | 82.0 | 80.4 | 97.8 | 90.0 | 93.8 | 94.0 | 94.0 | 94.0 |
| Class 12 | 88.2 | 62.5 | 73.2 | 92.3 | 100.0 | 96.0 | 95.7 | 91.7 | 93.6 |
| Class 13 | 68.6 | 68.6 | 68.6 | 90.6 | 94.1 | 92.3 | 80.4 | 80.4 | 80.4 |
| Accuracy (%) | 90.7 | | | 95.6 | | | 95.0 | | |

*4.2. Performance of Composite Models*

As mentioned previously, the number of axles is a key factor in the FHWA vehicle classification rules. Therefore, we generated the wheel positional features from the wheel locations detected by YOLOR and fed these features to the classifier together with the ViT encodings. To assess the benefits of adding the wheel positional features, we evaluated two model scenarios: (1) ViT models, which did not include wheel positional features, and (2) composite models, which included the ViT models as well as the wheel positional features from YOLOR. Table 4 shows the results of the ViT models (upper part) and composite models (bottom part).

Table 4. Comparison of the supervised ViT model and the composite models under different training schemes.

| Network | Top-1 Acc. (%) | Weighted Avg. Precision (%) | Weighted Avg. Recall (%) |
| --- | --- | --- | --- |
| ViT | 90.7 | 90.7 | 90.7 |
| ViT + DINO (freeze-encoder) | 94.6 | 94.6 | 94.6 |
| ViT + DINO | 95.6 | 95.7 | 95.6 |
| ViT + data2vec (freeze-encoder) | 93.5 | 93.5 | 93.5 |
| ViT + data2vec | 95.0 | 95.0 | 95.0 |
| ViT + YOLOR | 91.4 | 91.6 | 91.4 |
| ViT + DINO (freeze-encoder) + YOLOR | 95.4 | 95.5 | 95.4 |
| ViT + DINO + YOLOR | 96.0 | 96.0 | 96.0 |
| ViT + data2vec (freeze-encoder) + YOLOR | 95.0 | 95.0 | 95.0 |
| ViT + data2vec + YOLOR | 95.3 | 95.2 | 95.3 |

The composite models, which fused the ViT encodings and wheel positional features, improved the classification accuracy by 0.3–1.5%. This confirms that specific wheel positional features are important for the vehicle classification task. The DINO-pretrained ViT models were slightly better than their data2vec counterparts. The best composite model was "ViT + DINO + YOLOR", which achieved an overall accuracy of 96%. The detailed performance metrics (precision, recall, and F1-score) across classes are included in Table 5 for both scenarios: with and without the wheel features.

Table 5. Comparison of classification reports of the pretrained model (ViT + DINO) with and without wheel features.

|  | DINO (Pretrained) + YOLOR without wheel features | | | DINO (Pretrained) + YOLOR with wheel features | | |
| --- | --- | --- | --- | --- | --- | --- |
|  | Precision (%) | Recall (%) | F1-Score (%) | Precision (%) | Recall (%) | F1-Score (%) |
| Class 1 | 100.0 | 100.0 | 100.0 | 100.0 | 100.0 | 100.0 |
| Class 2 | 99.1 | 99.5 | 99.3 | 99.1 | 99.5 | 99.3 |
| Class 3 | 98.7 | 97.5 | 98.1 | 98.7 | 97.5 | 98.1 |
| Class 4 | 100.0 | 93.1 | 96.4 | 100.0 | 86.2 | 92.6 |
| Class 5 | 88.5 | 79.3 | 83.6 | 80.7 | 86.2 | 83.3 |
| Class 6 | 90.5 | 98.7 | 94.4 | 92.2 | 98.7 | 95.3 |
| Class 7 | 91.4 | 91.4 | 91.4 | 92.7 | 87.9 | 90.3 |
| Class 8 | 70.6 | 75.0 | 72.7 | 92.9 | 81.3 | 86.7 |
| Class 9 | 98.2 | 98.0 | 98.1 | 97.6 | 98.6 | 98.1 |
| Class 10 | 88.6 | 81.3 | 84.8 | 88.9 | 83.3 | 86.0 |
| Class 11 | 86.2 | 100.0 | 92.6 | 88.9 | 96.0 | 92.3 |
| Class 12 | 100.0 | 87.5 | 93.3 | 95.8 | 95.8 | 95.8 |
| Class 13 | 95.1 | 76.5 | 84.8 | 100.0 | 80.4 | 89.1 |
| Accuracy (%) | 96.3 | | | 96.6 | | |

The benefit of adding wheel features resulted in an obvious improvement in the F1-scores for classes 8 and 10. As indicated in Table 1, classes 8, 9, and 10 are all one-trailer trucks with minor differences (number of axles) and immensely imbalanced data distributions (with 82 images in class 8 but 2436 images in class 9). The model could have easily been confused among these classes. Adding the wheel positional features helped to better classify them, as well as classes 11, 12, and 13, which are all multi-trailer classes.

*4.3. Random Wheel Masking Strategy*

To regularize the learning process of the composite model, one of the YOLOR-detected wheels was randomly selected for masking when training the ViT. This could allow the learned ViT representations to adapt to the wheel noises being injected. The experiment results are summarized in Table 6.

Table 6. Comparison of model performance with and without random wheel masking.

| Network | Without Wheel Masking | | | Randomly Masking One Wheel | | |
|---|---|---|---|---|---|---|
| | Top-1 Acc. (%) | WAP * (%) | WAR * (%) | Top-1 Acc. (%) | WAP (%) | WAR (%) |
| ViT + YOLOR | 91.4 | 91.6 | 91.4 | 91.7 | 91.6 | 91.7 |
| ViT + DINO (freeze-encoder) + YOLOR | 95.4 | 95.5 | 95.4 | 96.0 | 96.0 | 96.0 |
| ViT + DINO + YOLOR | 96.3 | 96.3 | 96.3 | 96.7 | 96.8 | 96.7 |
| ViT + data2vec (freeze-encoder) +YOLOR | 95.0 | 95.0 | 95.0 | 96.5 | 96.6 | 96.5 |
| ViT + data2vec + YOLOR | 95.3 | 95.2 | 95.3 | 97.2 | 97.2 | 97.2 |

* WAP, WAR: abbreviations for weighted average precision and weighted average recall, respectively.

As shown in Table 6, all models had improved performance when one wheel was randomly masked during the training of the classifier. Without wheel masking, the best composite model was DINO + ViT + YOLOR, which achieved an accuracy of 96.3%. After applying one wheel masking, its accuracy was raised up to 96.7%. In contrast, the "ViT + data2vec + YOLOR" model benefited tremendously from the random wheel masking. Its accuracy was boosted by 1.9% to 97.2%, surpassing the DINO + ViT + YOLOR. Table 7 shows the detailed classification results of the ViT + data2vec + YOLOR for both the with and without wheel masking settings.

As indicated in Table 7, randomly masking one wheel increased the precision of classes 4, 5, 6, 7, 10, and 13 considerably. The F1-scores of most classes also improved, especially for the minority truck classes (5, 6, 7, 8, 10, and 13).

Table 7. Comparison of classification reports of the composite model (ViT + data2vec + YOLOR) with and without wheel masking.

|  | ViT + data2vec + YOLOR, without Wheel Masking | | | ViT + data2vec + YOLOR, with Wheel Masking | | |
| --- | --- | --- | --- | --- | --- | --- |
|  | Precision (%) | Recall (%) | F1-Score (%) | Precision (%) | Recall (%) | F1-Score (%) |
| Class 1 | 100.0 | 100.0 | 100.0 | 100.0 | 100.0 | 100.0 |
| Class 2 | 98.2 | 99.8 | 99.0 | 99.3 | 99.8 | 99.5 |
| Class 3 | 99.4 | 95.0 | 97.1 | 99.4 | 98.1 | 98.8 |
| Class 4 | 93.3 | 96.6 | 94.9 | 96.6 | 96.6 | 96.6 |
| Class 5 | 88.5 | 79.3 | 83.6 | 92.6 | 86.2 | 89.3 |
| Class 6 | 88.6 | 95.5 | 91.9 | 93.2 | 97.4 | 95.3 |
| Class 7 | 84.9 | 77.6 | 81.1 | 94.2 | 84.5 | 89.1 |
| Class 8 | 100.0 | 62.5 | 76.9 | 100.0 | 81.3 | 89.7 |
| Class 9 | 97.2 | 98.6 | 97.9 | 97.0 | 99.4 | 98.2 |
| Class 10 | 81.8 | 75.0 | 78.3 | 97.6 | 83.3 | 89.9 |
| Class 11 | 94.0 | 94.0 | 94.0 | 94.2 | 98.0 | 96.1 |
| Class 12 | 95.8 | 95.8 | 95.8 | 88.5 | 95.8 | 92.0 |
| Class 13 | 83.7 | 80.4 | 82.0 | 97.8 | 88.2 | 92.8 |
| Accuracy (%) | | 95.3 | | | 97.2 | |

## 5. Conclusions and Discussions

The two self-supervised learning methods (DINO and data2vec) showed their superiority over the supervised ViT. The classification accuracies were further boosted after applying DINO or data2vec for pretraining of the encoder. Finetuning the pretrained ViT encoder during the classifier training helped with the classification task. By adding additional wheel positional features, the models performed better than standalone ViTs. Additionally, the adoption of the random wheel masking strategy while finetuning the ViT encoder further improved the performance of the models, resulting in accuracies of 96.7 and 97.2%, respectively, for the DINO- and data2vec-pretrained models.

An important aspect to acknowledge is that classic supervised learning is largely constrained by limited annotated datasets. In contrast, self-supervised learning can take advantage of massive amounts of unlabeled data for representation learning and has become increasingly popular. Using self-supervised methods as a pretraining stage has been demonstrated to significantly improve the performance of vehicle classification. Between the two popular self-supervised learning methods, DINO and data2vec, there is an interesting finding: the DINO-pretrained ViT performed better than the data2vec-pretrained one, even with ViT finetuning and the addition of wheel positional features. However, by randomly masking a wheel during training, the data2vec-pretrained ViT outperformed the DINO-pretrained ViT. An arguable ratiocination is that during the pretraining stage, data2vec trains the ViT to predict the contextualized representations of masked image patches, which is consistent with our wheel masking strategy. This allows the data2vec-pretrained ViT to easily generalize over the masked wheel features, while for DINO, the ViT encoder learns from the cropped parts of input images and does not capture the contextual information, unlike with data2vec.

Although the ViT + data2vec + YOLOR model, coupled with the proposed strategy of random wheel masking, demonstrated excellent performance in classifying 13 FHWA vehicle classes, there is still plenty of room for future improvement. Dataset imbalance issues can be further mitigated by acquiring more images of minority class vehicles. YOLOR was adopted as a wheel detector to extract wheel positional features, which increases the computational footprint since the two standalone models (ViT and YOLOR) are executed in parallel. A unified model architecture could be investigated to reduce computational costs for practical real-time applications. The work presented in this paper implicitly assumes that the full bodies of all vehicles are visible in the images while this may not be true in real-world settings, where vehicle occlusion and superimposition often occur during heavy traffic conditions, causing only parts of vehicles to be visible. This issue could be mitigated by purposely training the models to recognize vehicle classes with partially blocked images. In fact, the data2vec method learns general representations by predicting contextualized latent representations of a masked view of the input in a self-distillation setting. Thus, data2vec-distilled representations are robust in cases of partial blocking of vehicles in images. Other mitigation methods may consider leveraging multiple views from different cameras or even multimodal sensory inputs. For example, using thermal cameras and LiDAR could help to improve model performance under low light conditions (e.g., at night).

**Acknowledgments:** The work presented in this paper is part of a research project (RP 20-04) sponsored by the Georgia Department of Transportation. The contents of this paper reflect the views of the authors, who are solely responsible for the facts and accuracy of the data, opinions, and conclusions presented herein. The contents may not reflect the views of the funding agency or other individuals.